\documentclass[10pt,twocolumn,letterpaper]{article}

\usepackage{cvpr}              

\usepackage[dvipsnames]{xcolor}
\usepackage{bm}
\usepackage{overpic}
\usepackage{enumitem} 
\usepackage{overpic} 
\usepackage{array}

\newcommand{\REMOVE}[1]{}

\newcommand{\methodname}{\textcolor{limegreen}{I}\textcolor{royalazure}{Re}\textcolor{redworst}{Ne}}

\definecolor{amethyst}{rgb}{0.6, 0.4, 0.8}
\definecolor{darkpastelgreen}{rgb}{0.01, 0.75, 0.24}
\definecolor{amber}{rgb}{1.0, 0.75, 0.0}
\definecolor{cadmiumorange}{rgb}{0.93, 0.53, 0.18}
\definecolor{lawngreen}{rgb}{0.49, 0.99, 0.0}
\definecolor{limegreen}{rgb}{0.2, 0.8, 0.2}
\definecolor{neongreen}{rgb}{0.22, 0.88, 0.08}
\definecolor{amethyst}{rgb}{0.6, 0.4, 0.8}
\definecolor{darkpastelgreen}{rgb}{0.01, 0.75, 0.24}
\definecolor{greenbest}{RGB}{88,137,15}
\definecolor{redworst}{rgb}{0.83, 0.3, 0.30}
\definecolor{royalazure}{rgb}{0.25, 0.41, 0.88}

\newcommand{\Alessiom}[1]{{\color{cyan} {[\bf am: #1]}}}

\newcommand{\Elenarmk}[1]{\textcolor{darkpastelgreen}{[\textbf{Elena}: {#1}]}}

\newcommand{\Adrian}[1]{\textcolor{amethyst}{[\textbf{Adrian}: {#1}]}}

\newcommand{\ferrmk}[1]{{\textcolor{royalazure} {[\bf Fer: #1]}}}

\newcommand{\MLP}{{\textsc{MLP}}} 
\newcommand{\MLPs}{{\textsc{MLP}_s}} 
\newcommand{\MLPc}{{\textsc{MLP}_c}} 

\newcommand{\x}{\textbf{x}}

\newcommand{\viewd}{\theta}
\newcommand{\density}{\sigma}
\newcommand{\co}{c}
\newcommand{\rgbedit}{\textrm{{I}}^{\textrm{edit}}}

\newcommand{\feat}{\it{f}}
\newcommand{\rco}{\MakeUppercase{c}}
\newcommand{\hiddenl}{\it{h}}
\newcommand{\penlayer}{\hiddenl}

\definecolor{cvprblue}{rgb}{0.21,0.49,0.74}
\usepackage[pagebackref,breaklinks,colorlinks,citecolor=cvprblue]{hyperref}

\usepackage{svg}
\usepackage{graphicx}
\usepackage{subcaption}
\usepackage{soul}
\usepackage{tabularx}
\usepackage{multirow}
\usepackage{comment}

\def\paperID{12486} 
\def\confName{CVPR}
\def\confYear{2024}

\begin{document}
\title{\textcolor{limegreen}{I}\textcolor{royalazure}{Re}\textcolor{redworst}{Ne}: \textcolor{limegreen}{I}nstant \textcolor{royalazure}{Re}coloring of \textcolor{redworst}{Ne}ural Radiance Fields}

\author{
\begin{tabular}{c c c}
Alessio Mazzucchelli$^{1,2}$ & Adrian Garcia-Garcia$^{1}$ & Elena Garces$^{3,5}$\\
Fernando Rivas-Manzaneque$^{4,8}$ & Francesc Moreno-Noguer$^{6}$ & Adrian Penate-Sanchez$^{7}$ \\
\end{tabular}
\\
\begin{tabular}{c c c }
\\
$^{1}$ Arquimea Research Center & $^{2}$ Universitat Politècnica de Catalunya & $^{3}$ Universidad Rey Juan Carlos \\
\end{tabular}
\\
\begin{tabular}{c c c}
$^{4}$ Volinga AI & $^{5}$ SEDDI & $^{6}$ Institut de Rob\`otica i Inform\`atica Industrial, CSIC-UPC  \\
\end{tabular}
\\
\begin{tabular}{c c}
$^{7}$Universidad de las Palmas de Gran Canaria, IUSIANI & $^{8} $Universidad Politécnica de Madrid
\end{tabular}
}



\twocolumn[{%
		\renewcommand\twocolumn[1][]{#1}%
		\maketitle
		\begin{center}
			\vspace{-0.65cm}	\hspace{-0.1cm}\includegraphics[width=1.0\linewidth, trim=0.5cm 0 0 0 ,clip]{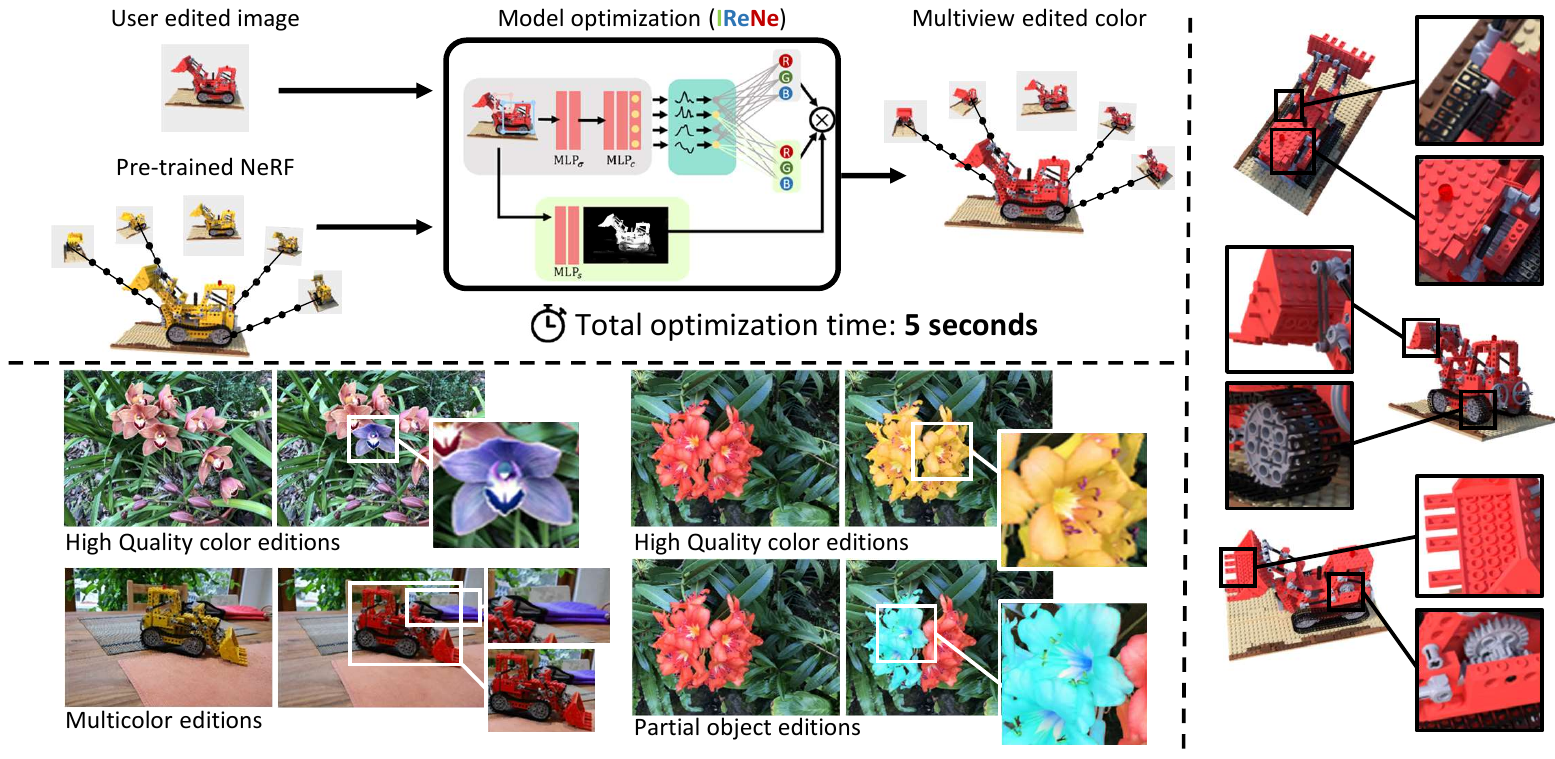}
   
		  \vspace{-3mm}
			\captionof{figure}{\methodname~enables instant 360$^\circ$ recoloring of pre-trained NeRFs using only a single image edited by the user (top row). We introduce an optimization scheme to avoid color bleeding at object boundaries and ensure consistency in view-dependent effects. Furthermore, as illustrated in the bottom row, various types of recoloring are possible, including full-object, partial-object, and multiple-object recoloring.
            }
		\label{fig:teaser}
	\end{center}
	}]
	
	\thispagestyle{empty}

\maketitle

\begin{abstract}
\vspace{-2mm}
 Advances in NERFs have allowed for 3D scene reconstructions and novel view synthesis. Yet, efficiently editing these representations while retaining photorealism is an emerging challenge. Recent methods face three primary limitations: they're slow for interactive use, lack precision at object boundaries, and struggle to ensure multi-view consistency.
%
We introduce \methodname~to address these limitations, enabling swift, near real-time color editing in NeRF. Leveraging a pre-trained NeRF model and a single training image with user-applied color edits, \methodname~swiftly adjusts network parameters in seconds. This adjustment allows the model to generate new scene views, accurately representing the color changes from the training image while also controlling object boundaries and view-specific effects.
%
Object boundary control is achieved by integrating a trainable segmentation module into the model. The process gains efficiency by retraining only the weights of the last network layer. 
We observed that neurons in this layer can be classified into those responsible for view-dependent appearance and those contributing to diffuse appearance. We introduce an automated classification approach to identify these neuron types and exclusively fine-tune the weights of the diffuse neurons. This further accelerates training and ensures consistent color edits across different views. A thorough validation on a new dataset, with edited object colors, shows significant quantitative and qualitative advancements over competitors, accelerating speeds by 5$\times$ to 500$\times$. 
\noindent The dataset is available on our project page: \href{https://iviazz97.github.io/irene/}{\texttt{iviazz97.github.io/irene}}.

\end{abstract}

\section{Introduction}
\label{sec:intro}

Neural Radiance Fields (NeRFs)~\cite{mildenhall2020nerf} have gained traction due to their ability to construct realistic 3D environments from 2D images and render high-fidelity, photorealistic novel viewpoints. Such advancements unlocked many possibilities, from immersive virtual environments~\cite{barron2023zipnerf,barron2022mipnerf360,rematas2022urf} to applications like augmented reality~\cite{qiao2019web,chen2023mobilenerf}. Nonetheless, the challenge of seamlessly and efficiently editing these neural representations while preserving photorealism remains a critical and largely unexplored frontier. 
Current NeRF color editing techniques~\cite{kuang2023palettenerf, Lee_2023_ICCV, gong2023recolornerf} face several limitations that impede their practical applications. First, total required time to perform edition is more than 1 minute for the fastest methods. This makes them ill-suited for interactive applications that demand real-time feedback. Second, current methods often lack accuracy, especially when managing color edits within object boundaries. Lastly, existing editing techniques encounter difficulties in sustaining consistent edits across various viewpoints, especially for scenes intended for 360-degree rendering.


In our paper, we present \methodname, a novel approach that facilitates the \underline{I}nstant \underline{Re}coloring of \underline{Ne}ural radiance fields, effectively addressing the challenges highlighted earlier. Leveraging an off-the-shelf, pre-trained NeRF model (in our case, Instant-NGP~\cite{mueller2022instant})
and a single training image featuring user-applied color adjustments to one or various objects, \methodname~swiftly fine-tunes the network parameters within seconds. This rapid fine-tuning enables the model to dynamically generate novel scene views, faithfully preserving the color modifications made in the training image. 

Our approach is founded on four pivotal contributions. Firstly, we augment the pre-trained NeRF model with a lightweight, trainable segmentation module, providing enhanced control over color edits within object boundaries. Secondly, speed is achieved by selectively fine-tuning the last layer of the network, leveraging only a single training image provided by the user. Thirdly, we show (and take advantage of) that neurons in this last layer show specialisation. Some neurons are responsible for rendering view-dependent appearance while others exclusively contribute to diffuse appearance. We implement an automated classification methodology to distinguish between these neuron types, enabling us to exclusively fine-tune the neurons associated with diffuse appearance. This strategy not only expedites the training process but also guarantees uniform color edits across various viewpoints, enhancing the model's consistency and performance.  


Up until now, there has been no dataset available for the quantitative assessment of NeRF color editing methodologies. The fourth contribution of our work involves the development of such a dataset, where the color of certain objects within the NeRF Synthetic~\cite{mildenhall2020nerf}, LLFF~\cite{mildenhall2019llff} and Mip-NeRF 360~\cite{barron2022mipnerf360} scenes have been meticulously edited manually using Photoshop. Through a comprehensive quantitative and qualitative evaluation on this dataset, it becomes evident that \methodname~outperforms~\cite{kuang2023palettenerf} by a significant margin. Moreover, it presents considerably better results than~\cite{Lee_2023_ICCV}, addressing issues such as color bleeding outside object boundaries and challenges in maintaining viewpoint consistency.  Most notably, our architecture's retraining can be accomplished in under 5 seconds—a stark contrast to the 1-minute requirement of~\cite{Lee_2023_ICCV} and the 30-120 minutes reconstruction time of~\cite{kuang2023palettenerf}. This capability opens the door to seamlessly integrating \methodname~into interactive image editing pipelines that demand immediate response. 

\begin{figure*}[t]
    \centering
    \includegraphics[width=1.0\textwidth]{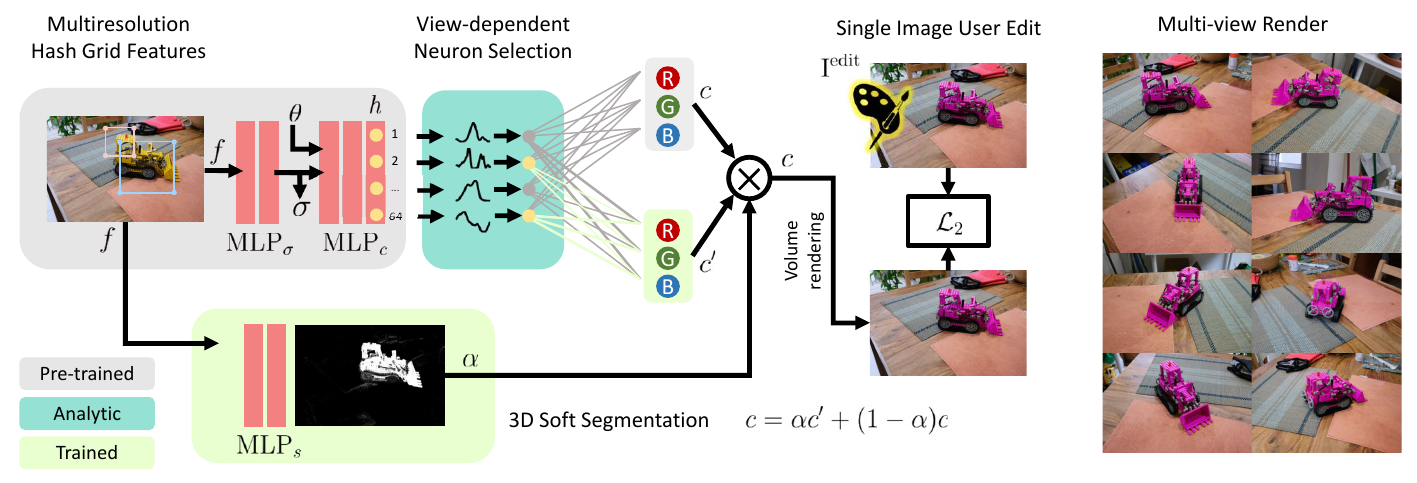}
\vspace{-5mm}
    \caption{{\bf Overview of \methodname.} We use a pre-trained NeRF and a user-edited training image $\rgbedit$. Our pre-trained NeRF is an Instant NGP~\cite{mueller2022instant} with: a density $\textrm{MLP}_\sigma$, with multiresolution hash features $f$, and a color $\textrm{MLP}_c$. Mapping the user's edits into the NeRF involves the following steps: 1) Automatic detection of the diffuse neurons in the last layer of $\textrm{MLP}_c$. 2) Training  an $\textrm{MLP}_s$, ruled by the features $f$, to estimate a volumetric soft-segmentation $\alpha$ of the edited region. 3) Fine-tuning the weights of the diffuse neurons in the last layer of $\textrm{MLP}_c$. 4) Alpha blending with the mask $\alpha$, to estimate the color of a 3D point $\x$ as a linear combination of the color $\co_{\x}$ predicted by the frozen weights with the color $\co^{\prime}_{\x}$ predicted by the retrained weights. 5) Volumetric rendering to obtain the edited image $\textrm{I}^\textrm{render}$. $\textrm{MLP}_s$ and the trainable last-layer neurons are optimized through standard RGB loss computation between $\textrm{I}^\textrm{render}$ and $\rgbedit$ in under 5 seconds.}
\vspace{-3mm}
    \label{fig:network}
    \centering
\end{figure*}
\section{Related Work}

Novel view synthesis of 3D scenes is a well-established field. Recently, Neural Radiance Fields (NeRF)~\cite{mildenhall2020nerf} marked a breakthrough in this domain. 
However, editing this new neural representation presents challenges because the 3D information is implicitly encoded in each neuron of a multi-layer perceptron (MLP).
Consequently, several works have aimed to extend the original NeRF formulation for different kinds of edits, such as scene relighting~\cite{NeRD, Neural_PIL, NeRFactor, NeRV}, scene composition~\cite{KunduCVPR2022PNF, liu2021editing, yang2021objectnerf}, object manipulation~\cite{jambon2023nerfshop}, inpainting~\cite{wang2023inpaintnerf360}, or text-based editing~\cite{wang2022clip,instructnerf2023,wang2023proteusnerf,wang2023nerf, wang2023seal3d}. 
In this work, we focus on the problem of editing the color of the NeRF, addressing the challenges associated with creating predictable, globally coherent, and fast edits.

\vspace{1mm}
\noindent\textbf{Color Editing.} Editing the color of an image has traditionally been done using two paradigms. Edit-propagation techniques, which involve using local colored cues, such as user scribbles or points, that are propagated to similar areas of the image~\cite{endo2016deepprop,an2008appprop,chen2012manifold,li2008scribbleboost}. 
The challenges lie in finding a similarity metric that can match the image features of the input exemplar and identify similar points in both local and non-local areas of the image. Deep learning-based approaches have been successful in addressing said challenges, leveraging the effectiveness of convolutional networks as feature extractors~\cite{meyer2018deep,zhang2019deep}.
The other paradigm is to use pre-computed color palettes ~\cite{chang2015palette,tan2018efficient,Du:2021:VRS}. These palettes contain the most relevant colors and serve as a basis for describing the rest through linear compositing. Editing palettes is a choice among artists who might combine it with object segmentation to perform global or local modifications. 

\vspace{1mm}
\noindent\textbf{NeRF Color Editing.} 
Directly applying color editing methods to a NeRF is not possible as the color values of the scene are implicitly stored in the neurons. A NeRF contains view-dependent effects such as specular reflections that need to be handled appropriately to avoid artifacts.
Some of the existing strategies focus on extending the original NeRF architecture so palette-based edits can be applied. 
RecolorNeRF~\cite{gong2023recolornerf} estimates a color palette in pixel space and, through optimization, decomposes each point into a weighted combination of color bases shared across the scene. Palette-NeRF~\cite{kuang2023palettenerf} improves it by accounting for specular (or view-dependent) effects. 
%
%
While producing compelling results, this method is very costly, taking hours to train, and is limited to global editions. Although, after said long training, multiple editions can be performed without additional computational cost.
Others, such as ICE-NeRF~\cite{Lee_2023_ICCV}, leverage user scribbles to propagate the edition to the full volume encoded by NeRF. In particular, ICE-NeRF utilizes user inputs to apply volume segmentation, which is later used for handling local color edits. 
As opposed to PaletteNeRF and RecolorNeRF, which optimize the full NeRF, ICE-NeRF fine-tunes only a specific set of neurons of a pre-trained color MLP, achieving faster convergence.  
However, while fast, its edits suffer from color bleeding at the borders. Furthermore, as acknowledged in the original work, the reprojection method utilized to propagate user inputs is not able to handle complex $360^{\circ}$ scenes.
Concurrently to us, ProteusNeRF~\cite{wang2023proteusnerf}, proposes a framework for interactive editing of NeRF, by propagating edits among views based on image features of a pre-trained model~\cite{caron2021emerging}. However, their results heavily rely on the view-consistency of these features, which are susceptible to failure under view-dependent effects such as specularities. Also, their edits take between 10 to 70 seconds in contrast to the 5 seconds of our work.

%

%

%

Inspired by these ideas, our method is also retrained from an existing NeRF. However, rather than a complete retraining, we selectively reuse information in the color MLP, akin to ICE-NeRF, expediting convergence. Additionally, we integrate a learnable segmentation branch capable of segmenting editable regions across images, even in sequences with $360^{\circ}$ views. As demonstrated in the results section, this segmentation, coupled with a selection and retraining of neurons responsible for diffuse rendering, effectively prevents color bleeding beyond the editable region and guarantees view-consistent edits.





\section{Background} \label{sec:background}



Many current NeRF methods use two modules. The first is a set of explicit trainable features
that encode 
each 3D point $\x$ in space in a high-dimensional feature vector $f_\x$. 
The second is a neural network, implemented as two multi-layer perceptrons, $\MLP_\density$ and $\MLP_\co$. 
$\MLP_\density$ takes as input the feature value for each point and outputs a density value, $\density$, and a hidden feature vector, $h$. 
To estimate the color of the given point, an encoding $\gamma(\cdot)$~\cite{mildenhall2020nerf,mueller2022instant} is applied to the view direction, $\viewd$, that, along with $h$, is fed into a second $\MLP_c$ that outputs the color $c$ of the point $\x$.
%
Each MLP is represented by a set of trainable parameters ($\phi_\density, \phi_\co$). Formally, 
%
%
\begin{align}
    \MLP_\density(\feat_{\x}, \phi_\density) &= \density_{\x}, h_{\x}  \label{eq:density_mlp} \\
    \MLP_\co(h_{\x},\gamma(\viewd), \phi_\co) &= \co_{\x, \viewd} \label{eq:color_mlp}
\end{align}  
%
%
%
Our method builds on a pre-trained NeRF using Instant-NGP~\cite{mueller2022instant}, which implements $\gamma$ using spherical harmonics. Another key characteristic of this method is that the learnable feature representation $\feat_\x$ is arranged into $L$ levels of a multiresolution hash grid where the number of features per level is determined as a geometric progression. At each level $l$ a hash grid function $\feat_{\x,l}$ produces the value of the features for the given resolution. The final feature is a concatenation of the features from each level:
\begin{equation} 
    \label{eq:multilevel_hashgrid}
        \feat_\x = \{ \feat_{\x, 0} \oplus \feat_{\x, 1} \oplus ... \oplus \feat_{\x, L} \}.
\end{equation}
The feature encodings $\{\feat_\x\}$ and the MLP weights for density and color ($\phi_\density$ and $\phi_\co$) are learned
at the same time in an end-to-end approach by minimizing a standard RGB loss between the volume rendered RGB value for each pixel $\rco_{ij}$ and the original RGB value from the training image $\rco'_{ij}$,
\begin{equation} 
    \label{eq:RGB_loss}
    \mathcal{L} = \sum{\| \rco_{ij} - \rco'_{ij} \|}_2^2
\end{equation}


\REMOVE{
\subsection{Last Layer}\Alessiom{Need to write it again and well}
We have opted to exclusively train the last hidden weights, the one between the final 64 neurons and the 3 RGB output. This decision is grounded in three key reasons:

Firstly, by narrowing our focus to fewer parameters, we expedite the training process. This not only enhances efficiency but also contributes to a more rapid convergence. \Alessiom{Full color MLPs params = 6336,
by training only last layer we touch 3percent of total params}

Secondly, our attempts to train a complete color Multi-Layer Perceptron (MLP) resulted in heightened overfitting, particularly on the same viewpoint. Moreover, it led to a loss of intricate details in unexplored regions of the image from that perspective
.\Alessiom{we can show the PSNR table of abalation and also show some images}

Thirdly, upon closer examination of neuron figures, we discovered a notable phenomenon in the last 64 neurons. Colors tend to automatically cluster within these neurons, implying that points in space sharing the same color are likely to activate identical neurons. This inherent clustering further justifies our decision to concentrate training efforts on these specific weights, optimizing the network for color representation.
\Alessiom{we show the images of the color clustering of the neurons}
}

\begin{figure}[t!]
  \centering
    \includegraphics[width=0.85\columnwidth]{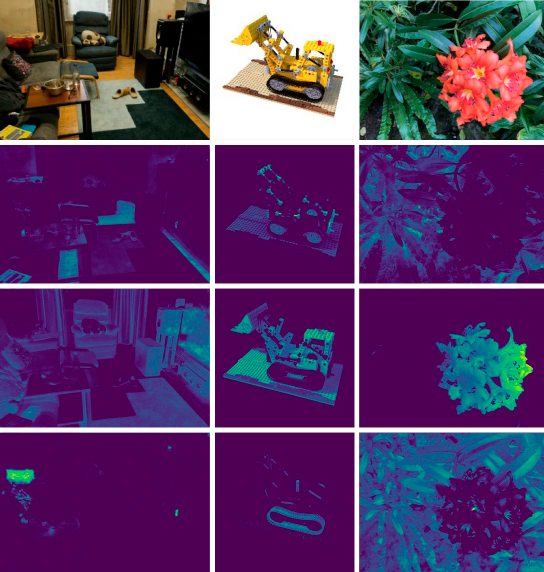}
\vspace{-2mm}
  \caption{Volumetric rendering for the activations of 3 neurons in the last layer. 
  Points with similar color in 3d space will share a similar activation pattern. }
\vspace{-3mm}
  \label{fig:activations}
\end{figure}

\section{Method}
    
Given as input a single re-colored view of the NeRF $\rgbedit \in \mathbb{R}^{W\times H \times3}$, our goal is to instantly propagate this edition to the entire NeRF taking into account the local or global nature of the edit and view-dependent effects. Addressing this requires overcoming several challenges: First, achieving near real-time NeRF editing is crucial. We accomplish this by selectively retraining only the last layer of the color MLP (refer to \cref{subsec:last_layer}).
This approach significantly reduces the number of parameters requiring modification compared to alternative methods~\cite{Lee_2023_ICCV, gong2023recolornerf, kuang2023palettenerf}. Secondly, since the edition is being performed only on a single view, being able to reproduce coherent view-dependent effects is inherently complex. To solve it, we propose to use the view-dependent information already present in the pre-trained NeRF (\cref{subsec:neuron_selection}). 
Finally, the NeRF edits need to be restricted to the modified regions. We prevent the edits from propagating to undesired areas by introducing a lightweight soft-segmentation network trained on the existing features (\cref{subsec:segmentation}). ~\cref{fig:network} presents an overview of the method.

\REMOVE{\subsection{Interactive 3D segmentation from a single } \label{sec:segmentation}

A challenge when editing a NeRF comes from determining which parts of the image correspond to the area being edited in other viewpoints.
This involves obtaining a 3D segmentation from the 2D information available in a single view. 
The classical approach would be to project the 2D points into a depth map, and then try to fill in the gaps by inferring the shape of the spatial envelope of the object. For instance,  ICE-NeRF~\cite{Lee_2023_ICCV} uses DIBR method~\cite{DIBR_paper} to obtain the 3D segmentation from a single depth map.
However, this is a sensible approach when dealing with 3D data as the appearance of a point in other views might differ greatly from the original 2D edited image. 
Our key idea is to take advantage of the implicit field given by the NeRFs voxel features~\cite{mueller2022instant}. Instead of traversing all the explicit 3D points as in other NeRF representations~\cite{mildenhall2020nerf}, we can directly access the multiresolution representation of each point given by the grid (Equation~\ref{eq:multilevel_hashgrid}).


We introduce a small $\MLPs$ that learns a soft-segmentation field $\alpha (\x) = \MLPs (\feat (\x))$ of the edited area using as input the features $\feat$ of the multiresolution hash grid. 
%
%
%
The network output is a single value that determines the probability $\alpha$ of a feature produced at a certain point of being part of the edited area. By using the features as what they are, a field, we can from a single view classify robustly which features belong to the edited part. 

\subsection{Color Remapping}\label{sec:color}

The output of the previous step is a volumetric segmentation of the points that likely belong to the surfaces \Alessiom{me and Elena liked better AREA instead of surfaces} being edited. 
Then, given this information, our next step is to actually recolor the NeRF given the user edits at the activations given by our segmentation mask $\alpha$. 

Because the output of the segmentation $\MLPs$ is a probability map that affects only some areas of the volume, we need to find a way to apply this segmentation to the color neurons. 
Our key idea is to clone the output color layer, obtaining $\co_\textrm{train}$ that defines the output values that have to be modified, and $\co_\textrm{freeze}$ that defines the values that need to remain the same according to the segmentation map. This can be modelled as applying an alpha blending to the output neurons as follows,
\begin{equation}
c (\x) = \left( \alpha (\x)  \co_\textrm{train} (\x) + (1- \alpha(\x))  \co_\textrm{freeze}(\x) \right)    
\end{equation}
Then, we train all our MLPs end-to-end using the RGB loss,
\begin{align} 
\mathcal{L} = \| \rgbedit - \textrm{I}^\textrm{render}  \|_2^2 \label{eq:segloss}
\end{align}
where $I^\textrm{render}$ is the image composed by volume rendering $c(\x)$

Adding this extra set of output neurons and only focusing on optimizing the new connections of this last layer have several advantages. First, it makes the process very efficient, as we only need to train a few weights of this last layer for every new edit. Second, it prevents overfitting, as our initial attempts to train a unique MLP resulted in heavily overfitting the input view. Finally, using this decoupling at the last layer we leverage an implicit color clustering that happens in this level of the architecture. \Elenarmk{We show the impact of these decisions in our ablation study in Section~\ref{sec:ablation}} \ferrmk{we don't, I think we should not include this parragraph (unless we have time enough to include it in the ablation study)}.

\REMOVE{Before we have explained how we can learn automatically to produce an object segmentation in the neural field that naturally works for any viewpoint from which we try to render. Once the segmentation is know our second contribution comes from how we propose to modify the NeRF model to be able to do colour remapping. In our work we have focused in being able to produce colour edition in a very fast manner that will enable the user to perform near interactive editions of the NeRF model. The proposed solution focuses on learning an additional final layer $\lambda'$ of the colour network $MLP_c$. We will then have the original final layer $\lambda$ and the final layer that contains the colour remapping $\lambda'$. Because our proposed approach on learns a single layer the time it takes to train the NeRF with the edition is of XXX seconds.

Performing a colour remapping is what most NeRF edition approaches like Palette-NeRF~\cite{Kuang_2023_CVPR} and ICE-NeRF~\cite{Lee_2023_ICCV} do. The main problem is that a colour remap has to be constrained to produce visually correct editions, ICE-NeRF~\cite{Lee_2023_ICCV} already identified in Fig 1 of their paper. ICE-NeRF also focused on only adjusting some of the layers of the NeRF, but the focused in trying to remap the internal layers of the network rather than just the final layer. We focus on only the final layer for two reasons. The first is that from our experiments modifying the internal layers of the network destroyed to much geometric information from the original scene. Because the last layer is a clear mapping from internal features to a RGB output we will show that our approach can handle colour editions by rewiring the connections of the last layer. And the second reason is that the only way to remain very fast to make possible interactive editions of the NeRF is by minimising the trainable parameters as much as possible (ICE-NeRF trains in 1 minute, we train in XXX seconds).

Our proposed approach trains an additional last layer $\lambda'$ and thus creates a modified $MLP'_c(\theta')$, were $\theta'$ is the same original weigths $\theta$ with the final layer $\lambda$ changed by the edited final layer $\lambda'$. The colour of each RGB pixel in the original model was determined by the following: $p_{i,j}=MLP_c(\theta,d)$. And the colour of the edited NeRF is established by $p'_{i,j}=MLP'_c(\theta',d)$. The final RGB value is then determined using the segmentation probability $\alpha$ as follows:

\begin{equation} 
    \label{eq:final_colour}
    P_{i,j}= \lambda * p'_{i,j} + (1-\lambda) * p_{i,j}
\end{equation}

The segmentation probability $\alpha$ and the edited final layer weights $\lambda'$ are trained together in a few cycles using the same RGB loss that we propagate along the neural network. This approach has showed to be very fast to train, very resilient to colour editions out of the object due to handling segmentation very robustly as we take advantage of the neural field properties.
}

\subsection{Identification of task decoupled neurons} \label{sec:viewdepend}

\Elenarmk{I don't like this title very much but I don't have a strong candidate,  maybe: Color Decoupling or View-dependent Decoupling? }

\Elenarmk{New from here (needs figure references, unify notation, and writing pass):}
In our final contribution we propose a novel way to handle view-dependent effects using our framework. 
After applying the color remapping process explained in the previous section, we end up having a globally consistent edition that works well in many cases but fails with some specular surfaces, in which we can observe color residuals, as shown in Figure~\ref{fig:ablation}~(e). 
This behavior makes sense as those values cannot be determined by editing just a single view, thus end up being unpredictable after the previous step.

Through several experiments we found out that the output neurons of the last layer of the color network $\MLPc$ specialize in two types of tasks: a set of neurons focus only on view-dependent effects, such as specularities or glossy reflections, while a different set of neurons capture the diffuse component of the surface.  Figure~\ref{fig:geocolor} illustrates this behavior. 
Even though our finding was empirical, we found that it applies to all scenes tested. 
We then propose a mechanism to identify which neurons have such behavior before training and freeze their values so that they produce the original color, as explained below.

For the points of the input edited view $\rgbedit$, we take the neuron activations of the last layer of the the color $\MLPc$ obtaining, $\hiddenl_\viewd \in \mathbb{R}^{W \times H \times 64 }$, for some view direction $\viewd$. 
%
%
Then, we uniformly sweep $\viewd$ along the azimuth every $\psi$ degrees obtaining a set of view-dependent feature vectors $ \{\hiddenl_{\viewd_0}, \hiddenl_{\viewd_1}, ..., \hiddenl_{\viewd_Q} \}$ where the value of $Q$ depends on the angular resolution of the scene as we discuss in the implementation details. 
We aggregate the values of this set along the spatial dimensions obtaining $A \in \mathbb{R}^{64 \times Q}$, where each $A_i \in \mathbb{R}^Q$ is a profile of the angular behavior of each output color neuron $i$ across the whole scene. \Elenarmk{make sure we unify the notation for the neuron $i$}

Then, we label the neuron $i$ as view-dependent neuron if one of two following conditions are satisfied: 1) if $\frac{\textrm{std} (A_i)}{ \mu (A_i)} < \tau_1$, where $\textrm{std}$ and $\mu$ are the mean and standard deviation, respectively; or 2) if $\min(A_i) \leq \tau_2$.
\Elenarmk{This needs plots of $A_i$}.
During training we freeze the weights of the neurons labeled as view-dependent which will maintain their original color.  
Our strategy of keeping the color of the view dependent effects unaltered aligns with related methods that explicitly isolate view-dependent effects from diffuse reflections using optimization~\cite{kuang2023palettenerf}. 
}

\subsection{Last Layer Re-training}
\label{subsec:last_layer}

Given a pre-trained NeRF model, the color $\MLP_\co$ can be seen as a mapping ruled by the output hidden vector $h_\x$ of the density $\MLP_\density$, by the view direction encoding $\gamma(\theta)$, and by the parameters $\phi_\co$ of the $\MLP_\co$, as seen in \cref{eq:color_mlp}. 
A na\"ive approach to perform the color edition would be to completely retrain the color MLP, obtaining a new set of parameters $\phi^{\prime}_c$ for the new set of colors $c^{\prime}_{x,\theta}$, as follows,
\begin{align}
    \MLP_\co(h_{\x},\gamma(\viewd), \phi^{\prime}_\co) &= \co^{\prime}_{\x, \viewd} \label{eq:new_color_mlp}.
\end{align}  

However, besides being a costly process due to the full retraining, this also leads to changes of important and useful information already available in the NeRF.
Thus, instead of re-training the full MLP, 
we propose to retrain only the \emph{last} \emph{layer} of the color $\MLPc$, formally, 
\begin{align}
    \MLP_{last}(\penlayer_{\x,\viewd}, \phi^{\prime}_{last}) &= \co^{\prime}_{\x, \viewd} \label{eq:last_layer},
\end{align}  
where $\penlayer_{\x,\viewd}$ is the output of the penultimate layer, that depends on the position $\x$ and the view direction $\viewd$, and $\phi^{\prime}_{last}$ is the new set of parameters for the last layer of the color $\MLPc$.
Although simple, this strategy enables us to achieve not only rapid edits but also, as demonstrated in \cref{subsec:ablation}, to generate higher-quality edits than full re-training. The updated color $\MLPc$ is obtained by exclusively retraining its last layer to match the single image user edit, using \cref{eq:RGB_loss}.
%
In Fig.~\ref{fig:activations} we show insights on why our approach performs so well without retraining the whole model. It can be seen that the neuron activations of the last layer show strong evidence of what seems to be color specialisation.

\begin{figure}[tb]
	\centering
	\includegraphics[width=1\linewidth]{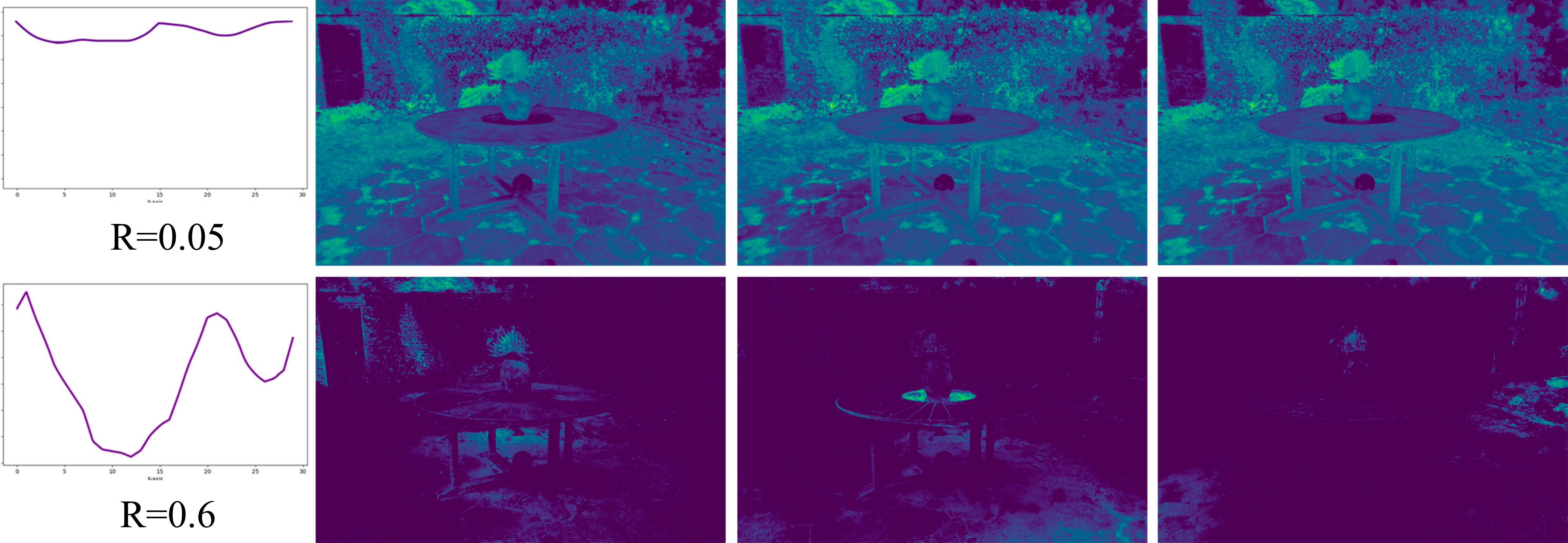}%
\vspace{-2mm}
	\caption{Rendered activations of the same pose while varying the view directional encoding. Top, diffuse neuron. Bottom, view-dependent neuron.  }
\vspace{-3mm}
	\label{fig:geocolor}
\end{figure}

\newcolumntype{Y}{>{\centering\arraybackslash}X}
\begin{table*} [h]
\begin{center}
{
\begin{tabularx}\textwidth{XlYYYYYYYYYYY}
\toprule
 & & \multicolumn{3}{c}{NeRF Synthetic~\cite{mildenhall2020nerf}} & \multicolumn{3}{c}{LLFF~\cite{mildenhall2019llff}} & \multicolumn{3}{c}{Mip NeRF 360~\cite{barron2022mipnerf360}} \\
\midrule
Method & &  PSNR$\uparrow$ & SSIM$\uparrow$ & LPIPS$\downarrow$ & PSNR$\uparrow$ & SSIM$\uparrow$ & LPIPS$\downarrow$ &  PSNR$\uparrow$ & SSIM$\uparrow$ & LPIPS$\downarrow$ \\
\midrule

\multicolumn{2}{l}{INGP (ref)~\cite{mueller2022instant}}         & 32.88 & 0.968 & 0.025 & 25.95 & 0.786 & 0.161 & 27.32 & 0.720 & 0.278 \\
\midrule
PaletteNeRF~\cite{kuang2023palettenerf} & & 29.92 & \textbf{0.960} & 0.038 & 24.02 & 0.763 & 0.200 & 22.24 & 0.666 & 0.351\\
RecolorNeRF~\cite{gong2023recolornerf}  & & 29.00 & 0.943 & 0.048 & 24.07 & 0.714 & 0.340 & - & - & -\\
\methodname                             & & \textbf{30.33} & {0.959} & \textbf{0.035} & \textbf{25.63} & \textbf{0.783} & \textbf{0.165} & \textbf{26.80} & \textbf{0.714} & \textbf{0.292}\\
\bottomrule
\end{tabularx}
}
\end{center}
\vspace{-4mm}
\caption{\small{\textbf{Quantitative results and comparison with  the state-of-the-art.} }}
\vspace{-3mm}
\label{table:main}
\end{table*}

\subsection{View-dependent Effects Preservation}
\label{subsec:neuron_selection}

The color $\MLPc$ contains all the information necessary to render the color of the scene, including view-dependent effects. Thus, by directly modifying $\phi_{last}$ using a single image to supervise the training, there is a high risk of modifying this information in an inconsistent manner, generating artifacts. 
To solve this, we propose a method to reuse the existing view-dependent information in the pre-trained NeRF.


Within the provided $\penlayer$, the 64-dimensional vector coming from the output of the penultimate layer, we identified 
two distinct behaviors: certain entries in this vector exhibit significant changes as the viewing direction changes, while others maintain constancy, encapsulating what we refer as geometry information (as shown in \cref{fig:geocolor}).
%
We preserve much of the view-dependent information by freezing part of the weights of this last layer.
By doing so, pre-existing information is retained after the edition process.

To achieve this, we propose a method to automatically identify entries in $\penlayer_k$ exhibiting such behavior. Specifically, for the pixels in the input edited view $\rgbedit$, we compute the neuron activations of the last layer, yielding  $\penlayer_\viewd \in \mathbb{R}^{W \times H \times 64 }$, for a given view direction $\viewd$. 
We then uniformly sweep $\viewd$ along the azimuth every $\psi$ degrees, obtaining a set of view-dependent feature vectors $ \{\penlayer_{\viewd_0}, \penlayer_{\viewd_1}, ..., \penlayer_{\viewd_Q} \}$ where the value of $Q$ depends on the angular resolution of the scene as we discuss in the implementation details. 
We aggregate the values of this set along the spatial dimensions obtaining $A \in \mathbb{R}^{64 \times Q}$, where each $A_k \in \mathbb{R}^Q$ is a profile of the angular behavior of each output color neuron $k$ across the whole scene (see \cref{fig:geocolor}). 

Subsequently, we label the neuron $k$ as a view-dependent neuron if either of the two conditions is met: 1) if $\frac{\textrm{std} (A_k)}{ \mu (A_k)} < \tau_1$, where $\textrm{std}(\cdot)$ and $\mu(\cdot)$ are the mean and standard deviation, respectively; or 2) if $\min(A_k) \leq \tau_2$.
During training, we preserve the weights of neurons identified as view-dependent, thereby retaining their original color.  
Our strategy of preserving the color of the view-dependent effects is consistent with related methods that explicitly separate view-dependent effects from diffuse reflections using optimization techniques~\cite{kuang2023palettenerf}.

\REMOVE{
\begin{figure*}[t]
  \centering
  \begin{subfigure}{\textwidth}
    \centering
    \includegraphics[width=0.2\linewidth]{figures/mip360/garden/vd_geo/0_plot.png}%
    \includegraphics[width=0.2\linewidth]{figures/mip360/garden/vd_geo/1.png}%
    \includegraphics[width=0.2\linewidth]{figures/mip360/garden/vd_geo/2.png}%
    \includegraphics[width=0.2\linewidth]{figures/mip360/garden/vd_geo/3.png}%
    \includegraphics[width=0.2\linewidth]{figures/mip360/garden/vd_geo/4.png}
  \end{subfigure}\\ 
  \begin{subfigure}{\textwidth}
    \centering
    \includegraphics[width=0.2\linewidth]{figures/mip360/garden/vd_geo/00_plot.png}%
    \includegraphics[width=0.2\linewidth]{figures/mip360/garden/vd_geo/22.png}%
    \includegraphics[width=0.2\linewidth]{figures/mip360/garden/vd_geo/333.png}%
    \includegraphics[width=0.2\linewidth]{figures/mip360/garden/vd_geo/44.png}%
    \includegraphics[width=0.2\linewidth]{figures/mip360/garden/vd_geo/55.png}
  \end{subfigure}
  
  \caption{Rendered activations of the same pose while varying the view directional encoding. Top, geometry neuron. Bottom, view-dependent neuron. \Elenarmk{candidate to be reduced. I will give a try} }
  \label{fig:geocolor}
\end{figure*}
}

\subsection{Soft 3D Segmentation from 2D User Edits}
\label{subsec:segmentation}
Since the information encoded on the color $\MLPc$ is global, performing modifications such as the one depicted in \cref{eq:last_layer} may lead to undesired edits in other regions of the scene.
To address this issue, we introduce a 3D segmentation step to selectively apply edits only in the target regions. Since our goal is to achieve near-real-time interactive editing, we avoid relying on  2D segmentation models~\cite{cen2023segment} or user scribbles~\cite{Lee_2023_ICCV} and the subsequent propagation of segmentation using depth information.
Instead, we propose to reuse the already existing information in the feature grid to implement a soft 3D segmentation model for our NeRF. We introduce a small $\MLPs$ that learns a soft-segmentation field $\alpha_\x = \MLPs (\feat_\x)$ for the edited area using as input the features $\feat$ of the multiresolution hash grid. The output of this network is a single value that represents the probability $\alpha$ of a feature at a specific point being part of the edited area.

We use this probability to perform alpha blending between the outputs of the network using the original set of parameters for the color $\MLP$ ($\phi_c$) and the new ones ($\phi^{\prime}_{last}$):
\begin{equation}
c_\x = \alpha_\x  \co^{\prime}_\x + (1- \alpha_\x)  \co_\x ,    
\end{equation}
where $\co^{\prime}_\x$ is the new color and $\co_\x$ is the original one.

The new $\MLPs$ introduced to perform the soft segmentation is trained alongside the last layer of the color $\MLPc$ using the loss presented in \cref{eq:RGB_loss}, without the need for any ground truth segmentation mask. Convergence is typically achieved in less than 5 seconds. 

\REMOVE{The final contribution of our work is what we have called identification of task decoupled neurons. In simple terms what this name entails is that through several experiments we found out that neurons in the last layer of the color network $MLP_c$ were specialising into two sets of tasks. 

-- \Adrian{Crear una figura que muestre experimentos de como las neuronas hacen unas view-dependent effects y otras geometry}

We show in Fig.~\ref{fig:neuron_specialisation} the behaviour of the neurons of the last layer. As it can be seen, some neurons focus only on performing view-dependent effects while others model the general geometry and colour of the scene. Through several experiments, we established this effect to be present in all dataset scenes. We began by freezing the layers by hand to see if 

What we then set to achieve once we discovered this affect to not just be an anomaly in a couple of scene was that we could take advantage of it if we could reliably detect the neurons that just performed view dependent effects.

-- \Adrian{Mostrar un ablation study de nuestro metodo con y sin la especialización de las neuronas}

We propose in this paper a way to determine which neurons are performing which task automatically to enhance the final results. Our approach naturally controls the view dependent effects that might affect the edited part of the object. What we need to establish is what neurons from the last layer $\lambda'$ of the edited color network $MLP'_c$ are performing exclusively view dependent effects. Once those neurons are detected we freeze them and not retrain them as part of $\lambda'$, the retain their original $\lambda$ values. So then $\lambda'$ is not a full set of final layer weights but rather a subset of the original $\lambda$ weights.

The process by which we detect neurons that only perform view-dependent effects is ........
}

\begin{figure*}[t]
  \centering
    \includegraphics[width=0.90\textwidth, trim={0 4mm 0 2mm}, clip]{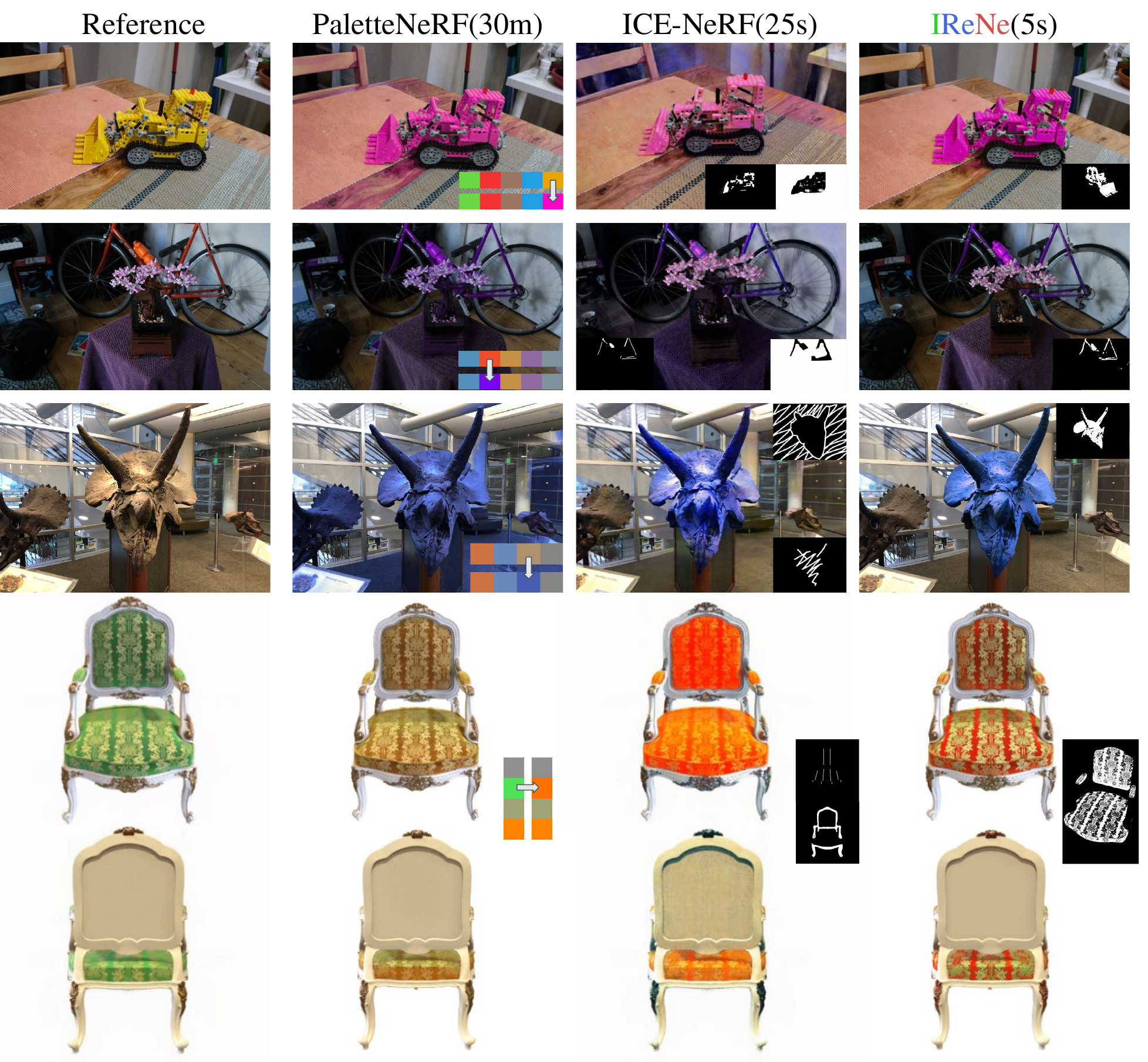}
\vspace{-2mm}
  \caption{{\bf Qualitative comparison with state of the art methods.} For each approach we show as a small overlayed image the input that the method requires from the user. For PaletteNeRF we show the original and the edited color palettes of the image. For \methodname, we show the region the user selects using Photoshop (or any similar editing tool). On that region we can interactively perform several color edits by modifying the HSV color within the region.}
\vspace{-3mm}
  \label{fig:recolors}
\end{figure*}

\begin{figure*}[t]
  \centering
    \includegraphics[width=1.0\textwidth, trim={9 62mm 8 0}, clip]{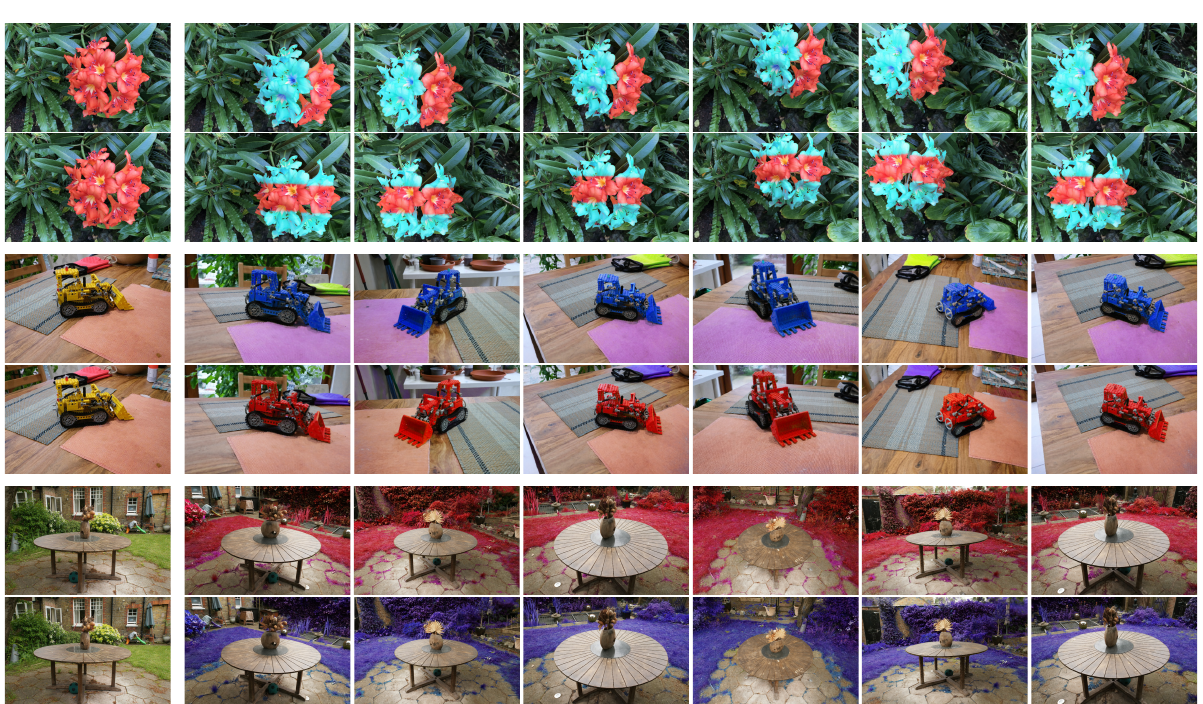}
\vspace{-5mm}
  \caption{ {\bf Additional qualitative results of ~\methodname.} First column: Original scene, Other columns: Edited scene. \textbf{Rows 1-2:} Two partial editions to the flower scene. \textbf{Rows 3-4:} Multicolor edition on the same scene. The third row has 3 edits: lego body (yellow to blue), lego emergency light (red to green), oven mitten (red to green). The fourth row has 2 edits. \textbf{Rows 5-6:} Single edit to recolor the vegetation.
  }
\vspace{-3mm}
  \label{fig:multicolor_edits}
\end{figure*}

\REMOVE{Needs to 
The segmentation $\MLPs$ has as an hidden layer of 64 neurons.
The new weights in the $\MLPc$ connecting $\co_\textrm{new}$ are loaded initially with the weights of $\co_\textrm{original}$. Same as PaletteNeRF we use the torch InstantNGP implementation \cite{torch-ngp}.
For our experiments environment we use PyTorch \cite{paszke2019pytorch}, and run on a single NVIDIA RTX
3090 GPU. We train every scene with 200 iterations over the same edited image using the Adam Optimizer \cite{kingma2014adam} with constant learning rate set to 0.01.
All gradient descent optimization of 

The soft segmentation $\MLPs$ is implemented as a two-layer MLP. The first layer takes as input the 32-dimensional vector representing the feature coming from the multiresolution hash grid an outputs a hidden feature of dimension 64. This hidden feature is fed into the output layer, which outputs the one-dimensional probability ($\alpha$). For the selection of $\epsilon$ entries, the values of $\tau_1$ and $\tau_2$ are 0.5 and 100 respectively. Q is equal to 30 for all the scenes. For Blender and MIP360 we sweep the whole $\psi$. Due to its forward-facing nature, only half of the $\psi$ is considered for LLFF. We used the the Pytorch~\cite{paszke2019pytorch} implementation of Instant-NGP~\cite{mueller2022instant} and we train our models on a single NVIDIA RTX 3090 GPU. We train every scene with 200 iterations over the same edited image using the Adam Optimizer \cite{kingma2014adam} with a constant learning rate set to 0.01.
}

\vspace{3mm}
\noindent{\bf Implementation details.}
The soft segmentation $\MLPs$ is implemented as a two-layer MLP. The input layer takes the 32-dimensional multiresolution hash grid feature vector $\feat$. The hidden layer has 64 neurons, the output is the one-dimensional probability ($\alpha$). Selected values of $\tau_1$ and $\tau_2$ are 0.5 and 100 respectively. Q is equal to 30 for all the scenes. For Blender and MIP360 we sweep the whole $\psi$. Due to its forward-facing nature, only half of the $\psi$ is considered for LLFF. We used the Pytorch~\cite{paszke2019pytorch} implementation of Instant-NGP~\cite{mueller2022instant}, our models were trained on a single NVIDIA RTX 3090 GPU. Every scene was trained for 200 iterations over the same edited image using the Adam Optimizer~\cite{kingma2014adam} with a constant learning rate set to 0.01.

\section{Evaluation}

In this section we describe how we achieved the first quantitative results of NeRF recoloring methods by creating ground truth edits that allow meaningful and detailed comparisons. We also show qualitative results to visually compare our method with the state of the art.

\vspace{3mm}
\noindent{\bf Dataset.} To the best of our knowledge, existing methods lack dedicated metrics to quantitatively evaluate the quality of the edition with respect to ground truth edits. We address this issue by creating a new benchmark building on existing datasets. From NeRF Synthetic~\cite{mildenhall2020nerf}, we used Lego, Mic, Drums, Ficus, and Hotdog (some texture files for Ship, Materials, and Chair are missing), and edited the color map before re-rendering the scene. We then used 1 view for training and 200 for evaluation. From LLFF~\cite{mildenhall2019llff} and MIP360~\cite{barron2022mipnerf360}, we used all scenes, for which we apply manually the same edits using Photoshop to 11 views (1 for training, 10 for evaluation). 



\subsection{Quantitative Evaluation}
We quantitatively compare our method with PaletteNeRF~\cite{kuang2023palettenerf} and RecolorNeRF~\cite{gong2023recolornerf}. Unfortunately, ICE-NeRF's code is not publicly available, so we just provide qualitative comparisons in the following subsection.
Both PaletteNeRF and RecolorNeRF are methods that require a color palette, while our dataset contains full re-colored images as input. 
Thus, for a fair comparison with these methods, we re-trained the models using our edited image in the RGB loss and froze all the parameters except the palette. In this way, we simulate the process of finding the most appropriate palette in an automated manner (applying stochastic gradient descent). As a result, we obtained the best palette for our edits and were able to emulate our full-image edits.

Quantitative evaluations are reported in Table~\ref{table:main}. As both \methodname~and PaletteNeRF are trained from Instant-NGP~\cite{torch-ngp}, we include metrics from this model on the ten evaluation images (without color edition) as a reference for the maximum achievable metric. Notably, \methodname~outperforms state-of-the-art methods across the three metrics—PSNR, SSIM, and LPIPS. Furthermore, on real datasets (LLFF and Mip NeRF 360), \methodname~achieves metrics comparable to Instant-NGP, suggesting proximity to the performance limit imposed by the pre-trained model. Results for RecolorNeRF on the Mip NeRF dataset are not reported due to this method not being able to converge on those scenes.
In terms of computation time, our model achieves remarkable speed-ups compared to other approaches. Specifically, the average training time for \methodname~is 5 seconds, while RecolorNeRF requires 15-20 minutes, and PaletteNeRF takes between 30 minutes and 2 hours. ICE-NeRF reports training times of 25 seconds. In summary, \methodname~achieves between 5$\times$ and 500$\times$ speed-ups compared to existing approaches. 

\begin{figure*}[t]
  \centering
  \includegraphics[width=1.0\textwidth]{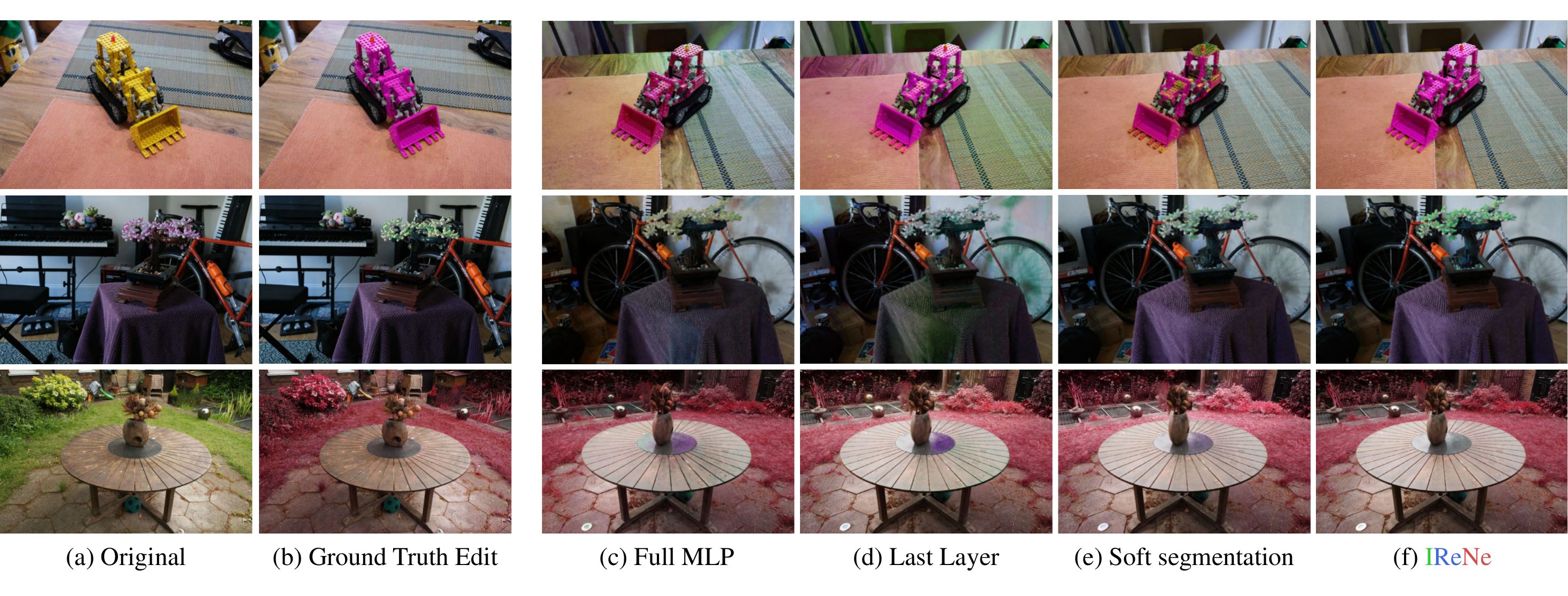}
\vspace{-8mm}
   \caption{\textbf{Qualitative ablation study.} Rendered images from the results detailed in Table~\ref{table:ablation}.}
\vspace{-0mm}
  \label{fig:ablation}
\end{figure*}

\newcolumntype{Y}{>{\centering\arraybackslash}X}
\begin{table*} [h]

\begin{center}
{
\begin{tabularx}\textwidth{XlYYYYYYYYYYYY}
\toprule
 & & & \multicolumn{3}{c}{NeRF Synthetic~\cite{mildenhall2020nerf}} & \multicolumn{3}{c}{LLFF~\cite{mildenhall2019llff}} & \multicolumn{3}{c}{Mip NeRF 360~\cite{barron2022mipnerf360}} \\
\midrule
Method & & &  PSNR$\uparrow$ & SSIM$\uparrow$ & LPIPS$\downarrow$ & PSNR$\uparrow$ & SSIM$\uparrow$ & LPIPS$\downarrow$ &  PSNR$\uparrow$ & SSIM$\uparrow$ & LPIPS$\downarrow$ \\
\midrule
\multicolumn{2}{l}{Full MLP}                & & 28.25 & 0.950 & 0.047 & 24.86 & 0.777 & 0.172 & 24.31 & 0.685 & 0.325\\
\multicolumn{2}{l}{Last layer}                & & 29.09 & 0.951 & 0.044 & 24.99 & 0.779 & 0.176 & 25.35 & 0.702 & 0.310\\
\multicolumn{2}{l}{Soft Segmentation}                   & & 30.20 & 0.956 & 0.037 & 24.77 & \textbf{0.784} & \textbf{0.163} & 25.68 & 0.707 & 0.295\\
\multicolumn{2}{l}{\methodname}          & & \textbf{30.33} & \textbf{0.959} & \textbf{0.035} & \textbf{25.63} & {0.783} & 0.165 & \textbf{26.80} & \textbf{0.714} & \textbf{0.292}\\
\bottomrule
\end{tabularx}
}
\end{center}
\vspace{-5mm}
    \caption{
    \small{\textbf{Ablation study.} In this ablation comparison we show the results of learning the full color MLP instead of just the last layer (Full MLP), the result of only training the last layer but not applying the soft segmentation MLP or the neuron selection (Last layer), the result of using the last layer and the soft segmentation (Soft Segmentation), and the full method \methodname.
    }
\vspace{-3mm}
}
\label{table:ablation}
\end{table*}



\subsection{Qualitative Evaluation}

We offer qualitative comparisons of our method with PaletteNeRF and ICE-NeRF. For PaletteNeRF, we manually selected the palette to achieve the most visually appealing image. Unfortunately, no code was provided for ICE-NeRF, so we had to use the result images from their paper and supplemental material. However, due to the low resolution of ICE-NeRF renders, we chose not to provide zoomed regions in the images to ensure a fair comparison.


\cref{fig:recolors} shows the comparison between different methods. Additionally, we present the original image and the supplementary information generated by each method (user scribbles for ICE-NeRF, color palette for PaletteNeRF, and the region selection that the user creates in Photoshop for \methodname~before applying the HSV color change in said region). 
The user edit that \methodname~uses is a region selection that the user does on one image, on which we then perform the desired color change in HSV space. 
%
%
For the first 2 results the camera pose of the render was not available for ICE-NeRF so we had to find by hand a similar pose and thus the slight misalignment between the pose used by PaletteNeRf and \methodname~and the pose in ICE-NeRFs results.

%


When analyzing the results in \cref{fig:recolors} it can be seen that ICE-NeRF performs poorly in the real scenes from the Mip NeRF 360 dataset, where the color bleeds out into the rest of the scene. Their results are better in Synthetic NeRF dataset but it can be seen that ICE-NeRF destroys part of the contained information when optimizing color. As a way to check if that degradation happens due to the method, the green chair images for the ground truth are obtained from the same ICE-NeRF source and serve as a quality baseline. In the LLFF dataset ICE-NeRF results are more robust. As PaletteNeRF is only able to support global recoloring, the edition is applied outside of the interest region (\emph{e.g.} Horns scene). It is also spilling some of the color edition into the view-dependent effects of other regions (\emph{e.g.} the pink color for the Lego is propagated to the yellow table reflections). 
%
We show additional results of our method in Fig.~\ref{fig:multicolor_edits}, with multicolor edits and other complex color editions.

\subsection{Ablation Study}
\label{subsec:ablation}

\cref{table:ablation} provides quantitative results on the quality of different variants of our method. Training uniquely the last layer of   $\MLPc$ results in better quality metrics than training the full MLP. Also, adding the soft segmentation further increases the quality of \methodname, while being able to retain existing view-dependent information in the pre-trained model leads to the highest quality. The effect of each of the contributions can be qualitative observed in \cref{fig:ablation}.

\section{Conclusions}
We introduced \methodname, a new method that enables interactive NeRF editing with a single user edit. The proposed method not only outperforms current state-of-the-art in terms of speed and interactivity but also overcomes the main limitations of the existing methods, namely multi-view consistency and precision at object boundaries.
Besides the high-quality results obtained by \methodname, we also provide insights on how color is encoded in NeRFs. We hope that this work may serve as inspiration to further lines of research, as well as enable use cases and edition workflows that were possible using traditional 2D/3D tools and, now, can be achieved in NeRFs thanks to \methodname.


\vspace{1mm}
\noindent{\bf Limitations and Future Work.}
One of the main limitations of our work is the need to rely on external edition tools, such as Photoshop, to achieve the complete edition. Also, while the last layer retraining and the method for discriminating which weights to freeze are very robust to the chosen image, the soft segmentation model results may be poor on some occasions. Besides trying to solve these limitations, future work should include the capacity to not only recolor regions but also being able to affect the indirect illumination produced by edited objects onto other objects. 
%

\section{Ackowledgements}
This work is supported by Arquimea Research Center and Horizon Europe, Teaming for Excellence, under grant agreement No 101059999, project QCircle, and by  project MoHuCo (PID2020-120049RB-I00). Elena Garces was partially supported by a Juan de la Cierva - Incorporacion Fellowship (IJC2020-044192-I).

{
    \small
    \bibliographystyle{ieeenat_fullname}
    \bibliography{main}

\begin{thebibliography}{39}
\providecommand{\natexlab}[1]{#1}
\providecommand{\url}[1]{\texttt{#1}}
\expandafter\ifx\csname urlstyle\endcsname\relax
  \providecommand{\doi}[1]{doi: #1}\else
  \providecommand{\doi}{doi: \begingroup \urlstyle{rm}\Url}\fi

\bibitem[li2(2008)]{li2008scribbleboost}
Scribbleboost: Adding classification to edge-aware interpolation of local image and video adjustments.
\newblock In \emph{Computer Graphics Forum}, pages 1255--1264. Wiley Online Library, 2008.

\bibitem[An and Pellacini(2008)]{an2008appprop}
Xiaobo An and Fabio Pellacini.
\newblock Appprop: all-pairs appearance-space edit propagation.
\newblock \emph{ACM Transactions on Graphics (TOG)}, 27\penalty0 (3):\penalty0 1--9, 2008.

\bibitem[Barron et~al.(2022)Barron, Mildenhall, Verbin, Srinivasan, and Hedman]{barron2022mipnerf360}
Jonathan~T. Barron, Ben Mildenhall, Dor Verbin, Pratul~P. Srinivasan, and Peter Hedman.
\newblock Mip-nerf 360: Unbounded anti-aliased neural radiance fields.
\newblock In \emph{Proceedings of the IEEE/CVF Conference on Computer Vision and Pattern Recognition (CVPR)}, pages 5470--5479, 2022.

\bibitem[Barron et~al.(2023)Barron, Mildenhall, Verbin, Srinivasan, and Hedman]{barron2023zipnerf}
Jonathan~T. Barron, Ben Mildenhall, Dor Verbin, Pratul~P. Srinivasan, and Peter Hedman.
\newblock Zip-nerf: Anti-aliased grid-based neural radiance fields.
\newblock In \emph{Proceedings of the IEEE/CVF International Conference on Computer Vision (ICCV)}, pages 19697--19705, 2023.

\bibitem[Boss et~al.(2021{\natexlab{a}})Boss, Braun, Jampani, Barron, Liu, and Lensch]{NeRD}
Mark Boss, Raphael Braun, Varun Jampani, Jonathan~T. Barron, Ce Liu, and Hendrik~P.A. Lensch.
\newblock Nerd: Neural reflectance decomposition from image collections.
\newblock In \emph{2021 IEEE/CVF International Conference on Computer Vision (ICCV)}, pages 12664--12674, 2021{\natexlab{a}}.

\bibitem[Boss et~al.(2021{\natexlab{b}})Boss, Jampani, Braun, Liu, Barron, and Lensch]{Neural_PIL}
Mark Boss, Varun Jampani, Raphael Braun, Ce Liu, Jonathan Barron, and Hendrik~PA Lensch.
\newblock Neural-pil: Neural pre-integrated lighting for reflectance decomposition.
\newblock In \emph{Advances in Neural Information Processing Systems}, pages 10691--10704, 2021{\natexlab{b}}.

\bibitem[Caron et~al.(2021)Caron, Touvron, Misra, J\'egou, Mairal, Bojanowski, and Joulin]{caron2021emerging}
Mathilde Caron, Hugo Touvron, Ishan Misra, Herv\'e J\'egou, Julien Mairal, Piotr Bojanowski, and Armand Joulin.
\newblock Emerging properties in self-supervised vision transformers.
\newblock In \emph{Proceedings of the IEEE/CVF International Conference on Computer Vision (ICCV)}, pages 9650--9660, 2021.

\bibitem[Cen et~al.(2023)Cen, Zhou, Fang, Yang, Shen, Xie, Jiang, Zhang, and Tian]{cen2023segment}
Jiazhong Cen, Zanwei Zhou, Jiemin Fang, Chen Yang, Wei Shen, Lingxi Xie, Dongsheng Jiang, Xiaopeng Zhang, and Qi Tian.
\newblock Segment anything in 3d with nerfs.
\newblock In \emph{NeurIPS}, 2023.

\bibitem[Chang et~al.(2015)Chang, Fried, Liu, DiVerdi, and Finkelstein]{chang2015palette}
Huiwen Chang, Ohad Fried, Yiming Liu, Stephen DiVerdi, and Adam Finkelstein.
\newblock Palette-based photo recoloring.
\newblock \emph{ACM Transactions on Graphics (TOG)}, 34\penalty0 (4):\penalty0 1--11, 2015.

\bibitem[Chen et~al.(2012)Chen, Zou, Zhao, and Tan]{chen2012manifold}
Xiaowu Chen, Dongqing Zou, Qinping Zhao, and Ping Tan.
\newblock Manifold preserving edit propagation.
\newblock \emph{ACM Transactions on Graphics (TOG)}, 31\penalty0 (6):\penalty0 1--7, 2012.

\bibitem[Chen et~al.(2023)Chen, Funkhouser, Hedman, and Tagliasacchi]{chen2023mobilenerf}
Zhiqin Chen, Thomas Funkhouser, Peter Hedman, and Andrea Tagliasacchi.
\newblock Mobilenerf: Exploiting the polygon rasterization pipeline for efficient neural field rendering on mobile architectures.
\newblock In \emph{Proceedings of the IEEE/CVF Conference on Computer Vision and Pattern Recognition (CVPR)}, pages 16569--16578, 2023.

\bibitem[Du et~al.(2021)Du, Lei, Xu, Tan, and Gingold]{Du:2021:VRS}
Zheng-Jun Du, Kai-Xiang Lei, Kun Xu, Jianchao Tan, and Yotam Gingold.
\newblock Video recoloring via spatial-temporal geometric palettes.
\newblock \emph{ACM Transactions on Graphics (TOG)}, 40\penalty0 (4), 2021.

\bibitem[Endo et~al.(2016)Endo, Iizuka, Kanamori, and Mitani]{endo2016deepprop}
Yuki Endo, Satoshi Iizuka, Yoshihiro Kanamori, and Jun Mitani.
\newblock Deepprop: Extracting deep features from a single image for edit propagation.
\newblock In \emph{Computer Graphics Forum}, pages 189--201. Wiley Online Library, 2016.

\bibitem[Gong et~al.(2023)Gong, Wang, Han, and Dou]{gong2023recolornerf}
Bingchen Gong, Yuehao Wang, Xiaoguang Han, and Qi Dou.
\newblock Recolornerf: Layer decomposed radiance fields for efficient color editing of 3d scenes.
\newblock \emph{arXiv preprint arXiv:2301.07958}, 2023.

\bibitem[Haque et~al.(2023)Haque, Tancik, Efros, Holynski, and Kanazawa]{instructnerf2023}
Ayaan Haque, Matthew Tancik, Alexei Efros, Aleksander Holynski, and Angjoo Kanazawa.
\newblock Instruct-nerf2nerf: Editing 3d scenes with instructions.
\newblock In \emph{Proceedings of the IEEE/CVF International Conference on Computer Vision (ICCV)}, 2023.

\bibitem[Jambon et~al.(2023)Jambon, Kerbl, Kopanas, Diolatzis, Drettakis, and Leimk{\"u}hler]{jambon2023nerfshop}
Cl{\'e}ment Jambon, Bernhard Kerbl, Georgios Kopanas, Stavros Diolatzis, George Drettakis, and Thomas Leimk{\"u}hler.
\newblock Nerfshop: Interactive editing of neural radiance fields.
\newblock \emph{Proceedings of the ACM on Computer Graphics and Interactive Techniques}, 6\penalty0 (1), 2023.

\bibitem[Kingma and Ba(2014)]{kingma2014adam}
Diederik~P Kingma and Jimmy Ba.
\newblock Adam: A method for stochastic optimization.
\newblock \emph{arXiv preprint arXiv:1412.6980}, 2014.

\bibitem[Kuang et~al.(2023)Kuang, Luan, Bi, Shu, Wetzstein, and Sunkavalli]{kuang2023palettenerf}
Zhengfei Kuang, Fujun Luan, Sai Bi, Zhixin Shu, Gordon Wetzstein, and Kalyan Sunkavalli.
\newblock Palettenerf: Palette-based appearance editing of neural radiance fields.
\newblock In \emph{Proceedings of the IEEE/CVF Conference on Computer Vision and Pattern Recognition}, pages 20691--20700, 2023.

\bibitem[Kundu et~al.(2022)Kundu, Genova, Yin, Fathi, Pantofaru, Guibas, Tagliasacchi, Dellaert, and Funkhouser]{KunduCVPR2022PNF}
Abhijit Kundu, Kyle Genova, Xiaoqi Yin, Alireza Fathi, Caroline Pantofaru, Leonidas Guibas, Andrea Tagliasacchi, Frank Dellaert, and Thomas Funkhouser.
\newblock {Panoptic Neural Fields: A Semantic Object-Aware Neural Scene Representation}.
\newblock In \emph{Proceedings of the IEEE/CVF Conference on Computer Vision and Pattern Recognition (CVPR)}, 2022.

\bibitem[Lee and Kim(2023)]{Lee_2023_ICCV}
Jae-Hyeok Lee and Dae-Shik Kim.
\newblock Ice-nerf: Interactive color editing of nerfs via decomposition-aware weight optimization.
\newblock In \emph{Proceedings of the IEEE/CVF International Conference on Computer Vision (ICCV)}, pages 3491--3501, 2023.

\bibitem[Liu et~al.(2021)Liu, Zhang, Zhang, Zhang, Zhu, and Russell]{liu2021editing}
Steven Liu, Xiuming Zhang, Zhoutong Zhang, Richard Zhang, Jun-Yan Zhu, and Bryan Russell.
\newblock Editing conditional radiance fields.
\newblock In \emph{Proceedings of the International Conference on Computer Vision (ICCV)}, 2021.

\bibitem[Meyer et~al.(2018)Meyer, Cornill{\`e}re, Djelouah, Schroers, and Gross]{meyer2018deep}
Simone Meyer, Victor Cornill{\`e}re, Abdelaziz Djelouah, Christopher Schroers, and Markus Gross.
\newblock Deep video color propagation.
\newblock In \emph{Proceedings of the British Machine Vision Conference (BMVC), Newcastle upon Tyne, UK, September 3-6, 2018}. British Machine Vision Association (BMVA), 2018.

\bibitem[Mildenhall et~al.(2019)Mildenhall, Srinivasan, Ortiz-Cayon, Kalantari, Ramamoorthi, Ng, and Kar]{mildenhall2019llff}
Ben Mildenhall, Pratul~P. Srinivasan, Rodrigo Ortiz-Cayon, Nima~Khademi Kalantari, Ravi Ramamoorthi, Ren Ng, and Abhishek Kar.
\newblock Local light field fusion: Practical view synthesis with prescriptive sampling guidelines.
\newblock \emph{ACM Transactions on Graphics (TOG)}, 2019.

\bibitem[Mildenhall et~al.(2020)Mildenhall, Srinivasan, Tancik, Barron, Ramamoorthi, and Ng]{mildenhall2020nerf}
Ben Mildenhall, Pratul~P. Srinivasan, Matthew Tancik, Jonathan~T. Barron, Ravi Ramamoorthi, and Ren Ng.
\newblock Nerf: Representing scenes as neural radiance fields for view synthesis.
\newblock In \emph{Proceedings of the European Conference on Computer Vision (ECCV)}, 2020.

\bibitem[M\"uller et~al.(2022)M\"uller, Evans, Schied, and Keller]{mueller2022instant}
Thomas M\"uller, Alex Evans, Christoph Schied, and Alexander Keller.
\newblock Instant neural graphics primitives with a multiresolution hash encoding.
\newblock \emph{ACM Trans. Graph.}, 41\penalty0 (4):\penalty0 102:1--102:15, 2022.

\bibitem[Paszke et~al.(2019)Paszke, Gross, Massa, Lerer, Bradbury, Chanan, Killeen, Lin, Gimelshein, Antiga, et~al.]{paszke2019pytorch}
Adam Paszke, Sam Gross, Francisco Massa, Adam Lerer, James Bradbury, Gregory Chanan, Trevor Killeen, Zeming Lin, Natalia Gimelshein, Luca Antiga, et~al.
\newblock Pytorch: An imperative style, high-performance deep learning library.
\newblock \emph{Advances in neural information processing systems}, 32, 2019.

\bibitem[Qiao et~al.(2019)Qiao, Ren, Dustdar, Liu, Ma, and Chen]{qiao2019web}
Xiuquan Qiao, Pei Ren, Schahram Dustdar, Ling Liu, Huadong Ma, and Junliang Chen.
\newblock Web ar: A promising future for mobile augmented reality—state of the art, challenges, and insights.
\newblock \emph{Proceedings of the IEEE}, 107\penalty0 (4):\penalty0 651--666, 2019.

\bibitem[Rematas et~al.(2022)Rematas, Liu, Srinivasan, Barron, Tagliasacchi, Funkhouser, and Ferrari]{rematas2022urf}
Konstantinos Rematas, Andrew Liu, Pratul~P. Srinivasan, Jonathan~T. Barron, Andrea Tagliasacchi, Tom Funkhouser, and Vittorio Ferrari.
\newblock Urban radiance fields.
\newblock \emph{Proceedings of the IEEE/CVF Conference on Computer Vision and Pattern Recognition (CVPR)}, 2022.

\bibitem[Srinivasan et~al.(2021)Srinivasan, Deng, Zhang, Tancik, Mildenhall, and Barron]{NeRV}
Pratul~P. Srinivasan, Boyang Deng, Xiuming Zhang, Matthew Tancik, Ben Mildenhall, and Jonathan~T. Barron.
\newblock Nerv: Neural reflectance and visibility fields for relighting and view synthesis.
\newblock In \emph{Proceedings of the IEEE/CVF Conference on Computer Vision and Pattern Recognition (CVPR)}, pages 7491--7500, 2021.

\bibitem[Tan et~al.(2018)Tan, Echevarria, and Gingold]{tan2018efficient}
Jianchao Tan, Jose Echevarria, and Yotam Gingold.
\newblock Efficient palette-based decomposition and recoloring of images via rgbxy-space geometry.
\newblock \emph{ACM Transactions on Graphics (TOG)}, 37\penalty0 (6):\penalty0 1--10, 2018.

\bibitem[Tang(2022)]{torch-ngp}
Jiaxiang Tang.
\newblock Torch-ngp: a pytorch implementation of instant-ngp, 2022.
\newblock https://github.com/ashawkey/torch-ngp.

\bibitem[Wang et~al.(2023{\natexlab{a}})Wang, Dutt, and Mitra]{wang2023proteusnerf}
Binglun Wang, Niladri~Shekhar Dutt, and Niloy~J Mitra.
\newblock Proteusnerf: Fast lightweight nerf editing using 3d-aware image context.
\newblock \emph{arXiv preprint arXiv:2310.09965}, 2023{\natexlab{a}}.

\bibitem[Wang et~al.(2022)Wang, Chai, He, Chen, and Liao]{wang2022clip}
Can Wang, Menglei Chai, Mingming He, Dongdong Chen, and Jing Liao.
\newblock Clip-nerf: Text-and-image driven manipulation of neural radiance fields.
\newblock In \emph{Proceedings of the IEEE/CVF Conference on Computer Vision and Pattern Recognition (CVPR)}, pages 3835--3844, 2022.

\bibitem[Wang et~al.(2023{\natexlab{b}})Wang, Jiang, Chai, He, Chen, and Liao]{wang2023nerf}
Can Wang, Ruixiang Jiang, Menglei Chai, Mingming He, Dongdong Chen, and Jing Liao.
\newblock Nerf-art: Text-driven neural radiance fields stylization.
\newblock \emph{IEEE Transactions on Visualization and Computer Graphics}, 2023{\natexlab{b}}.

\bibitem[Wang et~al.(2023{\natexlab{c}})Wang, Zhang, Abboud, and S{\"u}sstrunk]{wang2023inpaintnerf360}
Dongqing Wang, Tong Zhang, Alaa Abboud, and Sabine S{\"u}sstrunk.
\newblock Inpaintnerf360: Text-guided 3d inpainting on unbounded neural radiance fields.
\newblock \emph{arXiv preprint arXiv:2305.15094}, 2023{\natexlab{c}}.

\bibitem[Wang et~al.(2023{\natexlab{d}})Wang, Zhu, Ye, Huo, Ran, Zhong, and Chen]{wang2023seal3d}
Xiangyu Wang, Jingsen Zhu, Qi Ye, Yuchi Huo, Yunlong Ran, Zhihua Zhong, and Jiming Chen.
\newblock Seal-3d: Interactive pixel-level editing for neural radiance fields.
\newblock In \emph{Proceedings of the IEEE/CVF International Conference on Computer Vision (ICCV)}, pages 17683--17693, 2023{\natexlab{d}}.

\bibitem[Yang et~al.(2021)Yang, Zhang, Xu, Li, Zhou, Bao, Zhang, and Cui]{yang2021objectnerf}
Bangbang Yang, Yinda Zhang, Yinghao Xu, Yijin Li, Han Zhou, Hujun Bao, Guofeng Zhang, and Zhaopeng Cui.
\newblock Learning object-compositional neural radiance field for editable scene rendering.
\newblock In \emph{International Conference on Computer Vision ({ICCV})}, 2021.

\bibitem[Zhang et~al.(2019)Zhang, He, Liao, Sander, Yuan, Bermak, and Chen]{zhang2019deep}
Bo Zhang, Mingming He, Jing Liao, Pedro~V Sander, Lu Yuan, Amine Bermak, and Dong Chen.
\newblock Deep exemplar-based video colorization.
\newblock In \emph{Proceedings of the IEEE/CVF Conference on Computer Vision and Pattern Recognition (CVPR)}, pages 8052--8061, 2019.

\bibitem[Zhang et~al.(2021)Zhang, Srinivasan, Deng, Debevec, Freeman, and Barron]{NeRFactor}
Xiuming Zhang, Pratul~P. Srinivasan, Boyang Deng, Paul Debevec, William~T. Freeman, and Jonathan~T. Barron.
\newblock Nerfactor: Neural factorization of shape and reflectance under an unknown illumination.
\newblock \emph{ACM Trans. Graph.}, 40\penalty0 (6), 2021.

\end{thebibliography}
}


\end{document}